\documentclass[conference]{IEEEtran}
\IEEEoverridecommandlockouts
\usepackage{cite}
\usepackage{amsmath,amssymb,amsfonts}
\usepackage{algorithmic}
\usepackage{graphicx}
\usepackage{textcomp}
\usepackage{xcolor}

\usepackage{url}
\usepackage{graphicx}
\usepackage{epstopdf}

\usepackage{footnote}
\usepackage{balance}
\usepackage{amssymb}

\usepackage{amsmath}
\usepackage{multicol}
\usepackage{multirow}
\usepackage{anyfontsize}
\usepackage{array}
\usepackage{hhline}
\usepackage{float}
\usepackage{enumitem}
\usepackage{tabularx}

\def\BibTeX{{\rm B\kern-.05em{\sc i\kern-.025em b}\kern-.08em
    T\kern-.1667em\lower.7ex\hbox{E}\kern-.125emX}}
\begin{document}

\title{Automatic cinematography for 360$^\circ$ video}

\author{\IEEEauthorblockN{Hannes Fassold}
\IEEEauthorblockA{
\textit{JOANNEUM RESEARCH - DIGITAL}\\
Graz, Austria \\
hannes.fassold@joanneum.at}

}

\maketitle

\begin{abstract}
We describe our method for automatic generation of a visually interesting camera path (automatic cinematography) from a 360$^\circ$ video. Based on the information from the scene objects, multiple shot hypotheses for different shot types are constructed and the best one is rendered.
\end{abstract}


\section{Algorithm}

The goal of automatic cinematography is to calculate automatically a visually interesting camera path from a 360$^\circ$ video, in order to provide a traditional TV-like consumption experience. This is necessary for consumption of a 360$^\circ$ video on older TV sets, which do not provide any kind of interactive players for 360$^\circ$ video. Furthermore, even on devices capable of consuming 360$^\circ$ videos interactively, an user might prefer a lean-back mode, without the need to navigate around actively to explore the content. 

The proposed algorithm works in an iterative fashion, shot by shot. The shot length is set by the user, with a default of three seconds. Within the shot range, all objects (e.g. human, dog, cat, bicycle, car) appearing in the scene are detected and tracked throughout the shot with the deep learning based method proposed in \cite{Fassold2019Omnitrack}.
For each scene object, a set of measures is calculated, like the average size, average motion magnitude, \emph{neighbourhood score} (indicating how isolated the object is)  and \emph{visited
score} (indicating how visible the object was in the most recent shots). 

In the next phase, a set of \emph{shot hypotheses} is calculated. In order to match the diversity of traditional film which employs different shot types (from close up to very wide shots) for artistic purposes, we employ also several shot types. Specifically, we support the following shot types: tracking shot, static shot, medium shot, pan shot and recommender shot. Each shot type is aimed at a certain purpose, differs in how the saliency scores for the objects are calculated, and has a certain style in terms of artistic elements like FOV (field of view) and camera movement. For example, the tracking shot is focused on objects which are moving, large and isolated (like the singer of a band) and tracks them throughout the shot. On the other hand, for a static shot the camera is static and the focus is on a group of objects, like the audience in a concert. They differ also in their FOV, for the tracking shot we employ a standard FOV of 75$^\circ$, whereas for a static shot a wider FOV of 115$^\circ$ is used in order to capture groups better.

For each of these shot types, a couple of shot hypotheses (typically 2 – 4) are generated. For each shot hypothesis, a corresponding score is calculated which indicates how “good” this shot hypothesis is. The shot hypothesis score is influenced by a number of factors, like the saliency scores of the scene objects and rules of continuity editing (like the jump-cut rule for avoiding jump-cuts). The saliency score of an object indicates how "interesting" the object is and is dependent on several factors like shot type, object class, object size, motion magnitude, neighborhood score and visited score.

All generated shot hypotheses are collected in a set, and the shot hypothesis with the highest score is now
chosen as the best one. All frames of the current shot are now rendered, based on the viewport information provided in the best shot hypothesis. In order to not let a certain shot type dominate the generated video, we implemented a mechanism which limits the occurrence of each shot type in the generated video.

\begin{figure}[t]
	\centering
		\includegraphics[width=0.38\textwidth]{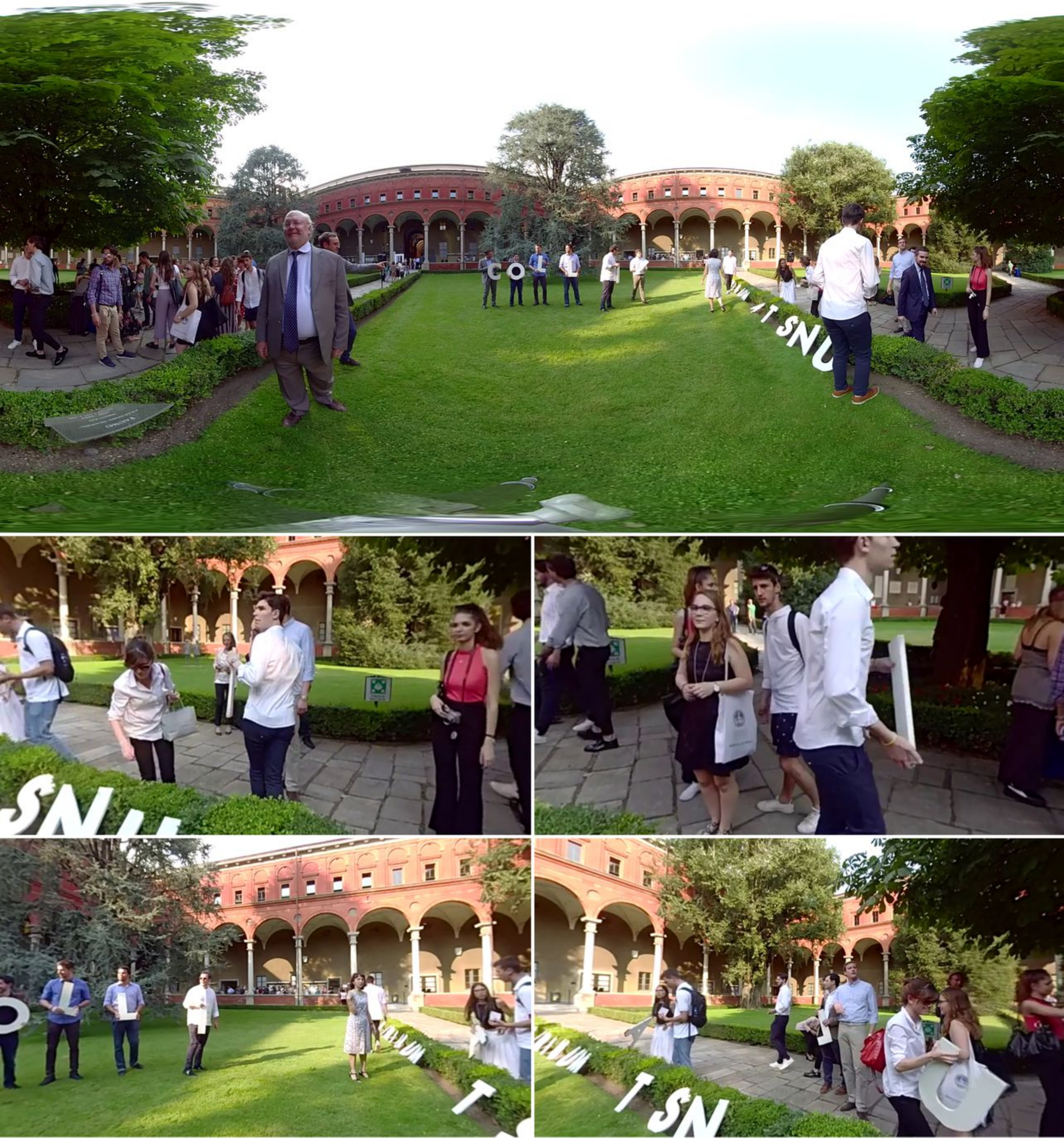}
	\caption{First row shows input video, second and third show the first two shots} 
	\label{fig:omniconnect}
\end{figure}

\section*{Acknowledgment}

This work has received funding from the European Union's Horizon 2020 research and innovation programme, grant n$^\circ$ 761934, Hyper360 (``Enriching 360 media with 3D storytelling and personalisation elements''). 


\bibliographystyle{IEEEtran}
\bibliography{references}

\end{document}